\documentclass[runningheads]{llncs}
\usepackage{makeidx}
\usepackage{graphicx}
\usepackage{amsmath,amssymb}
\usepackage{color}

\usepackage{multirow}

\begin{document}
	\pagestyle{headings}
	\mainmatter

	% Insert your submission number here
	\def\GCPR18SubNumber{0133}

	% Replace with your title
	\title{Decoupling Respiratory and Angular Variation\\ in Rotational X-ray Scans Using a\\ Prior Bilinear Model}

	\titlerunning{Decoupling Respiratory and Angular Variation in Rotational X-ray}
	\authorrunning{Geimer et al.}
	\author{	Tobias~Geimer\inst{1,2,3} \and
				Paul~Keall\inst{4} \and
				Katharina~Breininger\inst{1} \and
				Vincent~Caillet\inst{4} \and
				Michelle~Dunbar\inst{4} \and
				Christoph~Bert\inst{2,3} \and
				Andreas~Maier\inst{1,2}
	}
	\institute{Pattern Recognition Lab, Department of Computer Science, Friedrich-Alexander-Universit\"at Erlangen-N\"urnberg, Germany \and
		Erlangen Graduate School in Advanced Optical Technologies (SAOT), Friedrich-Alexander-Universit\"at Erlangen-N\"urnberg, Germany \and
		Department of Radiation Oncology, Universit\"atsklinikum Erlangen, Friedrich-Alexander-Universit\"at Erlangen-N\"urnberg, Germany \and
		ACRF Image X Institute, The University of Sydney, Australia\\ \email{tobias.geimer@fau.de}}

	\maketitle

	\begin{abstract}
		Data-driven respiratory signal extraction from rotational X-ray scans is a challenge as angular effects overlap with respiration-induced change in the scene.
		In this paper, we use the linearity of the X-ray transform to propose a bilinear model based on a prior 4D scan to separate angular and respiratory variation.
		The bilinear estimation process is supported by a B-spline interpolation using prior knowledge about the trajectory angle. Consequently, extraction of respiratory features simplifies to a linear problem. Though the need for a prior 4D CT seems steep, our proposed use-case of driving a respiratory motion model in radiation therapy usually meets this requirement.
		We evaluate on DRRs of 5 patient 4D CTs in a leave-one-phase-out manner and achieve a mean estimation error of $3.01\,\%$ in the gray values for unseen viewing angles. We further demonstrate suitability of the extracted weights to drive a motion model for treatments with a continuously rotating gantry.
		%\keywords{Bilinear Model \and Motion Model \and Respiratory Signal \and X-ray Projection \and Feature Extraction}
	\end{abstract}

	\section{Introduction}
	\label{sec:intro}
	Extracting information about a patient's breathing state from X-ray projections is important for many time-resolved application, such as CT reconstruction or motion compensation in image-guided radiation therapy (IGRT)~\cite{Keall2006}.
	In the context of CT imaging, respiratory motion during the scan time introduces data inconsistencies that, ultimately, manifest in reconstruction artifacts. These effects can be mitigated by incorporating non-linear motion models into state-of-the-art algorithms for motion-compensated reconstruction~\cite{McClelland2013}.
	In IGRT, the consequences of respiratory motion may be particularly harmful if not addressed properly.
	Here, malignant tumor cells are irradiated following an optimized dose distribution that is the result of treatment planning based on CT imaging.
	However, respiratory motion may lead to a displacement of the target volume during irradiation, resulting in underdosage of the tumor and, ultimately, the potential survival of malignant cells~\cite{Keall2006}.
	Motion tracking represents the state-of-the-art procedure, where the treatment beam is continuously following the tumor motion. However, this process requires sophisticated motion monitoring, often incorporating motion models to estimate internal deformation from the available imaging modalities such as on-board X-ray imagers.
	
	Many data-driven approaches have been proposed to extract respiratory information from X-ray projections, ranging from the established Amsterdam Shroud \cite{Yan2013} to more sophisticated approaches based on epipolar consistency~\cite{Aichert2015}.
	In the context of respiratory motion models~\cite{McClelland2013} used to estimate internal deformation fields in IGRT~\cite{Geimer17b}, their main drawback is the fact that most of these methods only extract a 1D signal, that, at best, can be decomposed into amplitude and phase~\cite{Fassi2015}. When the motion representation is covered by a statistical shape model (SSM)~\cite{Cootes1995,McClelland2013}, it is highly desirable to extract more features to allow for a lower reconstruction error.
	While approaches exist that extract multiple respiratory features from X-ray projections~\cite{Fischer2017}, they are restricted to static acquisition angles. As training a separate model for every possible static angle is infeasible, these methods are typically not suited for applications with a continuously rotating gantry, such as cone-beam CT or volumetric arc therapy.
	
	With both angular and respiratory variation present in the X-ray images of a rotational scan, we aim to separate these effects by expressing them as the two domains of a bilinear model with corresponding rotational and respiratory feature space. A mathematical foundation to decompose multiple sources of variation was given by De Lathauwer \textit{et al.}~\cite{DeLathauwer2000} who formulated Higher-order Singular Value Decomposition (HOSVD) on a tensor of arbitrary dimensionality. Meanwhile, bilinear models have seen use in many fields including medical applications. Among others, Tenenbaum \textit{et al.}~\cite{Tenenbaum2000} used a bilinear model to separate pose and identity from face images, while \c{C}imen \textit{et al.}~\cite{Cimen2014} constructed a spatio-temporal model of coronary artery centerlines.
	
	In this work, we show that due to the linearity of the X-ray transform, a bilinear decomposition of respiratory and angular variation exists (Sec.\ \ref{sec:radon}).
	Subsequently, Sec.\ \ref{sec:training} and \ref{sec:estimation} will cover both model training and its application in feature estimation.
	With the simultaneous estimation of both bilinear weights being an ill-posed problem, we propose a B-spline interpolation of rotational weights based on prior knowledge about the trajectory angle. With known rotational weights, the task of estimating respiratory weights reduces to a linear one. We validate our model in a leave-one-phase-out manner using digitally reconstructed radiographs (DRRs) of five patient 4D CTs consisting of eight respiratory phases each. The main purpose of our evaluation is to give a proof-of-concept of the proposed decoupling process. In addition, we provide first indication that the extracted respiratory features contain volumetric information suitable for driving a motion model.
	
	\section{Material \& Methods}
	\label{sec:method}
	
	\subsection{X-ray Transform under Respiratory and Angular Variation}
	\label{sec:radon}
	An X-ray projection $\boldsymbol{p}_{i,j}\in \mathbb{R}^{N^2 \times 1}$ at rotation angle $\phi_i \in \left[0,2\pi\right)$ and respiratory phase $t_j \in \left[0,1\right)$ is given by the X-ray transform $\boldsymbol{R}_i\in \mathbb{R}^{N^2\times N^3}$ applied to the volume $\boldsymbol{v}_j \in \mathbb{R}^{N^3\times 1}$
	
	\begin{equation}
		\boldsymbol{p}_{i,j} = \boldsymbol{R}_i \, \boldsymbol{v}_j,
		\label{equ:radon}
	\end{equation}
	where N indicates the arbitrary dimension of the volume and projection image.
	It has been shown that the respiratory-induced changes in the anatomy can be described by an active shape model of the internal anatomy~\cite{McClelland2013}:
	\begin{equation}
		\boldsymbol{v}_j = \boldsymbol{M} \, \boldsymbol{a}_j + \boldsymbol{\bar{v}},
		\label{equ:linear}
	\end{equation}
	where $\boldsymbol{\bar{v}}$ is the data mean of the mode,  $\boldsymbol{M} \in \mathbb{R}^{N^3 \times f}$ contains the eigenvectors corresponding to the first $f$ principal components, and $\boldsymbol{a}_j \in \mathbb{R}^{f \times 1}$ are the model weights corresponding to phase $t_j$.
	Thus, Eq.\ \ref{equ:radon} can be further processed to
	\begin{eqnarray}
		\boldsymbol{p}_{i,j} &=& \boldsymbol{R}_i \, \left(\boldsymbol{M} \, \boldsymbol{a}_j + \boldsymbol{\bar{v}}\right) \nonumber \\
							% &=& \boldsymbol{R}_i \, \boldsymbol{M} \, \boldsymbol{a}_j + \boldsymbol{R}_i \, \boldsymbol{\bar{v}} \nonumber \\ 
							 &=& \boldsymbol{M}^{R}_i  \, \boldsymbol{a}_j + \boldsymbol{\bar{p}}_i.
		\label{equ:radon2}
	\end{eqnarray}
	Now, $\boldsymbol{M}^{R}_i \in \mathbb{R}^{N^2 \times f}$ represents a model (with mean $\boldsymbol{\bar{p}}_i = \boldsymbol{R}_i \, \boldsymbol{\bar{v}}$) for describing the projection image $\boldsymbol{p}_{i,j}$ under the fixed angle $\phi_i$ given the respiratory weights $\boldsymbol{a}_j$.
	The inversion of the X-ray transform $\boldsymbol{R}$ itself is ill-posed for a single projection. However, $\boldsymbol{M}^R_i$ can be inverted more easily with $\boldsymbol{a}_j$ encoding mostly variation in superior-inferior direction observable in the projection. As a result, respiratory model weights can be estimated from a single projection image if the angle-dependent model matrix is known~\cite{Fischer2017}.
	
	Furthermore, we propose a X-ray transform $\boldsymbol{R}_i$ to be approximated by a linear combination using a new basis of X-ray transforms, such that
	\begin{equation}
		\boldsymbol{R}_i = \sum\limits_{k}b_{k,i} \boldsymbol{R}_k.
	\end{equation}
	This essentially mimics Eq.\ \ref{equ:linear} with $\boldsymbol{R}_k$ describing variation in the projection images solely caused by the rotation of the gantry. The resulting scalar factors $b_{k,i}$ form the weight vector $\boldsymbol{b}_i = \left[\dots, b_{k,i}, \dots \right]$.
	Note that in this formulation, we implicitly assume a continuous trajectory and that the breathing motion is observable from each view. This gives rise to a bilinear formulation for any given projection image $\boldsymbol{p}_{i,j}$. However, bilinear models typically do not operate on mean-normalized data. Therefore, we use the decomposition described in Eq.\ \ref{equ:linear} without mean subtraction:
	
	\begin{eqnarray}
	\boldsymbol{p}_{i,j} 	&=& \boldsymbol{R}_i \, \left( \boldsymbol{M} \, \boldsymbol{a}_j \right) \nonumber \\ 
							&=& \sum\limits_{k}b_{k,i} \, \boldsymbol{R}_k \left( \boldsymbol{M} \, \boldsymbol{a}_j \right) \nonumber \\
							&=& \sum\limits_{k}b_{k,i} \, \boldsymbol{M}^R_k \, \boldsymbol{a}_j  \nonumber \\
							&=& \mathcal{M} \times_1 \boldsymbol{a}_j \times_2 \boldsymbol{b}_i, \label{equ:bilinear}
	\end{eqnarray}
	
	\noindent
	where $\mathcal{M} \in \mathbb{R}^{N^2 \times f \times g}$ is a model tensor with respiratory and rotational feature dimensionality $f$ and $g$.
	Here, $\times_\ast$ denotes the mode product along the given mode $\ast$. For more details on tensor notation please refer to~\cite{Kolda2009}.
	
	\subsection{Model Training}
	\label{sec:training}
\begin{figure}[tb]
		\centering
		\begin{minipage}{0.48\textwidth}
			\centering
			\includegraphics[width=0.95\textwidth]{./graphics/math/TensorUnfoldingD}
		\end{minipage}
		\vline
		\begin{minipage}{0.48\textwidth}
			\centering
			\includegraphics[width=0.95\textwidth]{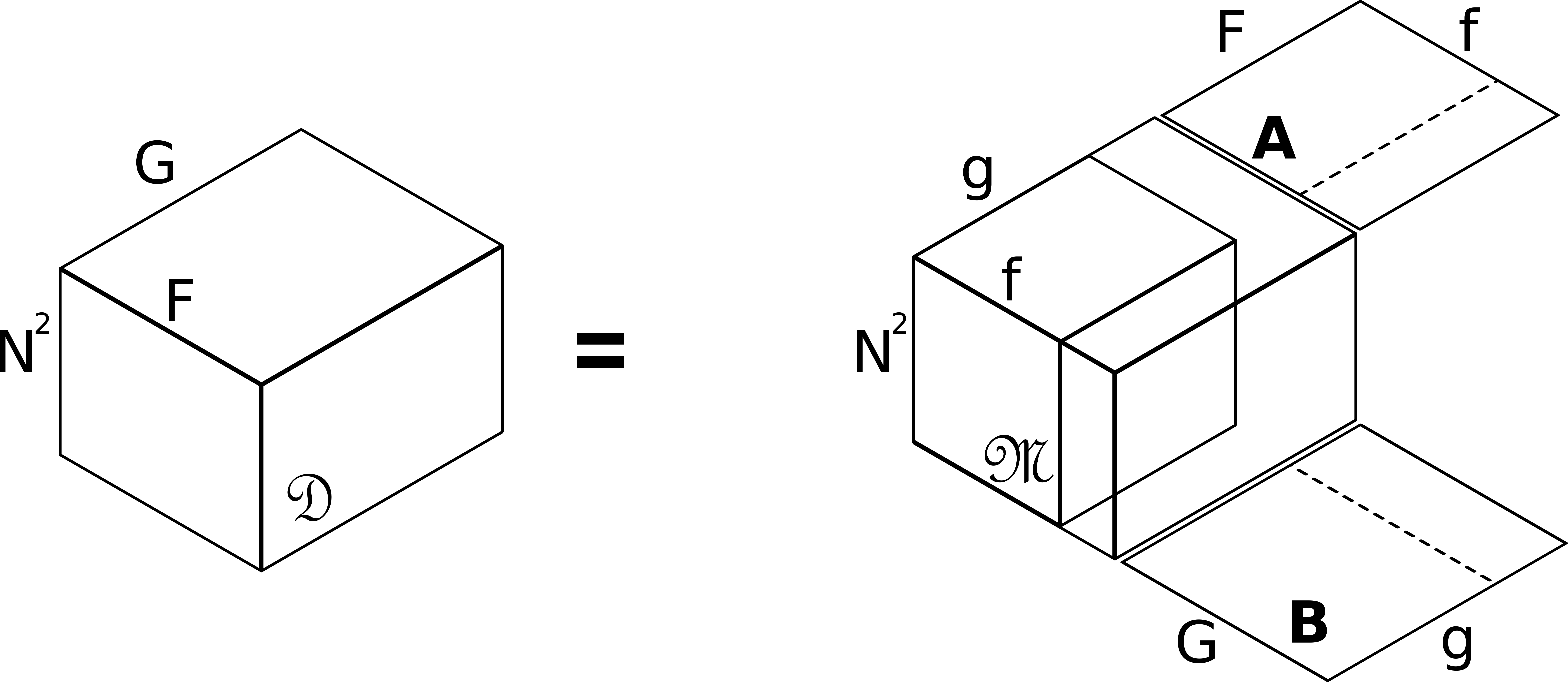}
		\end{minipage}
		\caption{\textbf{Left}: Tensor unfolding concatenates the slices of a tensor along a selected mode. \textbf{Right}: Dimensionality reduction (HOSVD) of the rotational and respiratory domain.}
		\label{fig:unfoldBilinear}
	\end{figure}

	For model training, a prior 4D CT scan is required yielding $F$ phase-binned volumes $\boldsymbol{v}_j, \, j \in \left\{1,\dots,F\right\}$.
	Using the CONRAD software framework~\cite{Maier2013}, DRRs are computed at $G$ angles $\phi_i, \, i \in \left\{1,\dots,G\right\}$ along a circular trajectory. The resulting $F\cdot G$ projection images $\boldsymbol{p}_{i,j}$ form the data tensor $\mathcal{D} \in \mathbb{R}^{N^2 \times F \times G}$.
	Using HOSVD~\cite{DeLathauwer2000} we perform dimensionality reduction on the data tensor.
	First, $\mathcal{D}$ is unfolded along mode $k$:
	\begin{equation*}
		\mathcal{D}_{(k)} \in \mathbb{R}^{d_k \times (\prod\limits_{l\neq k} d_l)}.
	\end{equation*}
	Fig.\ \ref{fig:unfoldBilinear}-left illustrates the unfolding process.
	Second, SVD is performed on each unfolded matrix
	\begin{equation}
		\mathcal{D}_{(k)} = \boldsymbol{U}_k \boldsymbol{S}_k \boldsymbol{V}_k^\top,
	\end{equation}
	which yields the tensor basis  $\{\boldsymbol{U}_k\}_{k=1}^K$ to project $\mathcal{D}$ onto (see Fig.\ \ref{fig:unfoldBilinear}-right):
	\begin{equation}
		\mathcal{D} = \mathcal{M} \times_{k=1}^K \boldsymbol{U}_k.
	\end{equation}
	
	\noindent
	Finally, $\mathcal{D}$ can be described by a model tensor $\mathcal{M}$:
	\begin{equation}
	\underbrace{\mathcal{D}}_{N^2 \times F \times G} = \underbrace{\mathcal{M}}_{N^2 \times f \times g} \times_1 \underbrace{\boldsymbol{A}}_{F \times f} \times_2 \underbrace{\boldsymbol{B}}_{G \times g},
	\end{equation}
	where $\boldsymbol{A}$ and $\boldsymbol{B}$ carry the low-dimensional ($f\ll F,\, g\ll G$) model weights for respiratory and angular variation, respectively.

	\subsection{Weight Estimation}
	\label{sec:estimation}
	Given an observed projection image $\boldsymbol{p}_{x,y}$ at unknown respiratory phase $t_y$ and angle $\phi_x$, our objective is to find coefficients $\boldsymbol{a}_y$, $\boldsymbol{b}_x$ for respiration and rotation to best represent the observation in terms of the model:
	\begin{equation}
		\boldsymbol{p_{x,y}} = \mathcal{M} \times_1 \boldsymbol{a}_y \times_2 \boldsymbol{b}_x.
	\end{equation}
	However, as $\boldsymbol{a}_y$ and $\boldsymbol{b}_x$ need to be optimized simultaneously, this task is highly ill-posed. %degenerates into a Chicken-\&-Egg problem.
	Tenenbaum \textit{et al.}~\cite{Tenenbaum2000} used an Expectation-Maximization algorithm to cope with this problem in their separation of identity and pose in face images. However, they benefit from the fact that only the pose is a continuous variable whereas identity is a discrete state, drastically simplifying their EM-approach. In our case, both respiratory and angular variation have to be considered continuous.
	Fortunately, we can incorporate prior knowledge about the trajectory into the estimation process. From the trajectory, the angle of each projection image is known even though the corresponding weights $\boldsymbol{b}_x$ are only given for particular angles within the training samples. Consequently, interpolating the desired rotational weights using those within the training set seems feasible under the assumption of a continuous non-sparse trajectory.
	
	\subsubsection{Rotational B-spline Interpolation.}
	\label{sec:bspline}
	Using the rotation angle as prior knowledge, we propose extending the bilinear model with a B-spline curve fitted to the rotational weights $\boldsymbol{b}_i$ from training:
	\begin{equation}
		\boldsymbol{b}(u) = \sum\limits_{i=1}^{G} \boldsymbol{b}_i \mathcal{N}_i(u),
	\end{equation}
	with $\mathcal{N}_i$ being the B-spline basis functions.
	Using a uniform parametrization with respect to the training angles, $u(\phi)$ of new angle $\phi$ is given as
	\begin{equation}
		u(\phi) = \frac{\phi - \phi_\text{min}}{\phi_\text{max} - \phi_\text{min}}.
	\end{equation}
	
	\subsubsection{Respiratory Weight Computation.}
	\label{sec:respiratory}
	With the rotational weight interpolated, multiplying $\mathcal{M}$ and $\boldsymbol{b}\left(u\left(\phi_x\right)\right)$, first removes the angular variation:
	\begin{equation}
		\mathcal{M}^R_x = \mathcal{M} \times_2 \boldsymbol{b}\left(u\left(\phi_x\right)\right) \in \mathbb{R}^{N^2 \times f \times 1}.
	\end{equation}
	Collapsing the 1-dimension results in the angle-dependent model matrix $\boldsymbol{M}^R_x \in \mathbb{R}^{N^2 \times f}$ for the new angle $\phi_x$.
	Closing the loop to Eq.\ \ref{equ:radon2} without mean, computation of the respiratory weights simplifies to a linear problem solved via the pseudo-inverse of $\boldsymbol{M}^R_x$:
	\begin{equation}
		\boldsymbol{a}_y = \left(\boldsymbol{M}^R_x\right)^{-1} \, \boldsymbol{p}_{x,y}.
	\end{equation}	
	
	For application, the extracted respiratory weights could either be used as input for a respiratory model~\cite{Geimer17b} to estimate internal deformation fields, or for data augmentation in the context of cone-beam CT by generating additional gated projections for different rotational weights $\boldsymbol{b}_i$ at the constant phase corresponding to $\boldsymbol{a}_x$ (see Eq.\ \ref{equ:bilinear}).
	
	\subsection{Data \& Experiments}
	\label{sec:evaluation}
	Evaluation was performed on 4D CTs of five patients, consisting of eight phase-binned volumes each at respiratory states $0\,\%$, $15\,\%$, $50\,\%$, $85\,\%$, $100\,\%$ inhale, and $85\,\%$, $50\,\%$, $15\,\%$ exhale.
	Using CONRAD~\cite{Maier2013}, DRRs of size $512\times 512$ with an $0.8\,\text{mm}$ isotropic pixel spacing were created by forward projecting each of the $F=8$ volumes at $G=60$ angles over a circular trajectory of $360^\circ$ (in $6^\circ$ steps). Consequently, the full data tensor $\mathcal{D}$ featured a dimensionality of $512^2 \times 8 \times 60$ for each patient. With the model being patient-specific, the following experiments were conducted individually for each patient and results were averaged where indicated.
	
	\paragraph{Experiment 1.} Our goal is to provide a proof-of-concept of the bilinear decoupling and to investigate how accurately respiratory weights can be extracted using the proposed method.
	For each patient, a dense bilinear model was trained on the entire data tensor to assess the variance explained by the respiratory domain. For comparison, a SSM (linear PCA with mean normalization) of the 4D CT volumes was trained to assess weights and variance prior to the influence of the X-ray transform.
	
	\paragraph{Experiment 2.} To investigate how well a previously unseen projection image for unknown angles and breathing phase can be decomposed into rotational and respiratory weights, every 6th angle (10 in total) was removed from training. Further, leave-one-out evaluation was performed, were each phase was subsequently removed prior to training, resulting in a sparser data tensor of size $512^2 \times 7 \times 50$.
	In this scenario, the dense bilinear model provided a reference for the features to be expected. In the evaluation step, the corresponding projection image and its respective trajectory angle were fed to the model and the rotational and respiratory weights were estimated as described in section \ref{sec:estimation}. From these weights, the projection image was rebuilt and model accuracy was assessed with respect to the mean gray-value error between the reconstructed image and the original.
	
	\paragraph{Experiment 3.} To assess their use for predicting 3D information, the bilinear respiratory weights were used as a driving surrogate for a motion model~\cite{McClelland2013}, which, apart from the surrogate, usually consists of an internal motion representation and a internal-external correlation model. For motion representation, a SSM was trained for each test phase on the remaining seven 4D CT volumes (Eq.\ \ref{equ:linear}).
	With the 4D SSM weights and bilinear respiratory weights showing similar behavior but differing in scale (Fig.\ \ref{fig:weightsAndVariance}), multi-linear regression~\cite{Wilms2014a} was chosen to correlate the bilinear weights to the 4D SSM weights. True to the leave-one-out nature, the regression matrix $\boldsymbol{W} \in \mathbb{R}^{e\times f}$ was trained between the weights of the seven remaining phases relating bilinear respiratory weights of feature dimension $f=6$ to 4D SSM weights of dimension $e=5$ (see also results of experiment 1). The rebuilt 3D volume was then compared to the ground truth volume in the 4D CT in terms of HU difference.
	
	\section{Results \& Discussion}
	\label{sec:results}
	\begin{figure}[tb]
		\centering
		\centerline{
			\includegraphics[width=0.45\textwidth]{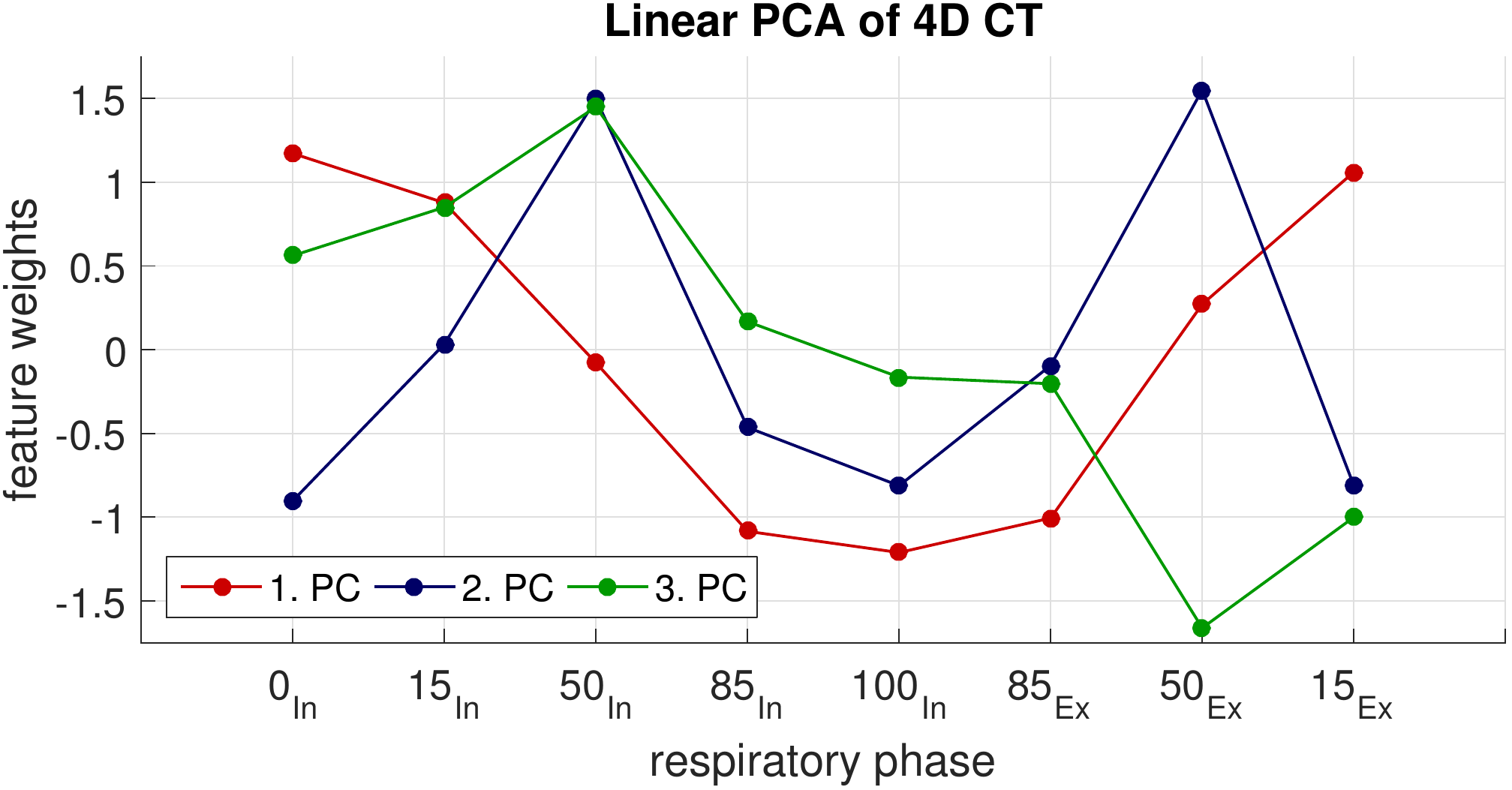}
			\includegraphics[width=0.45\textwidth]{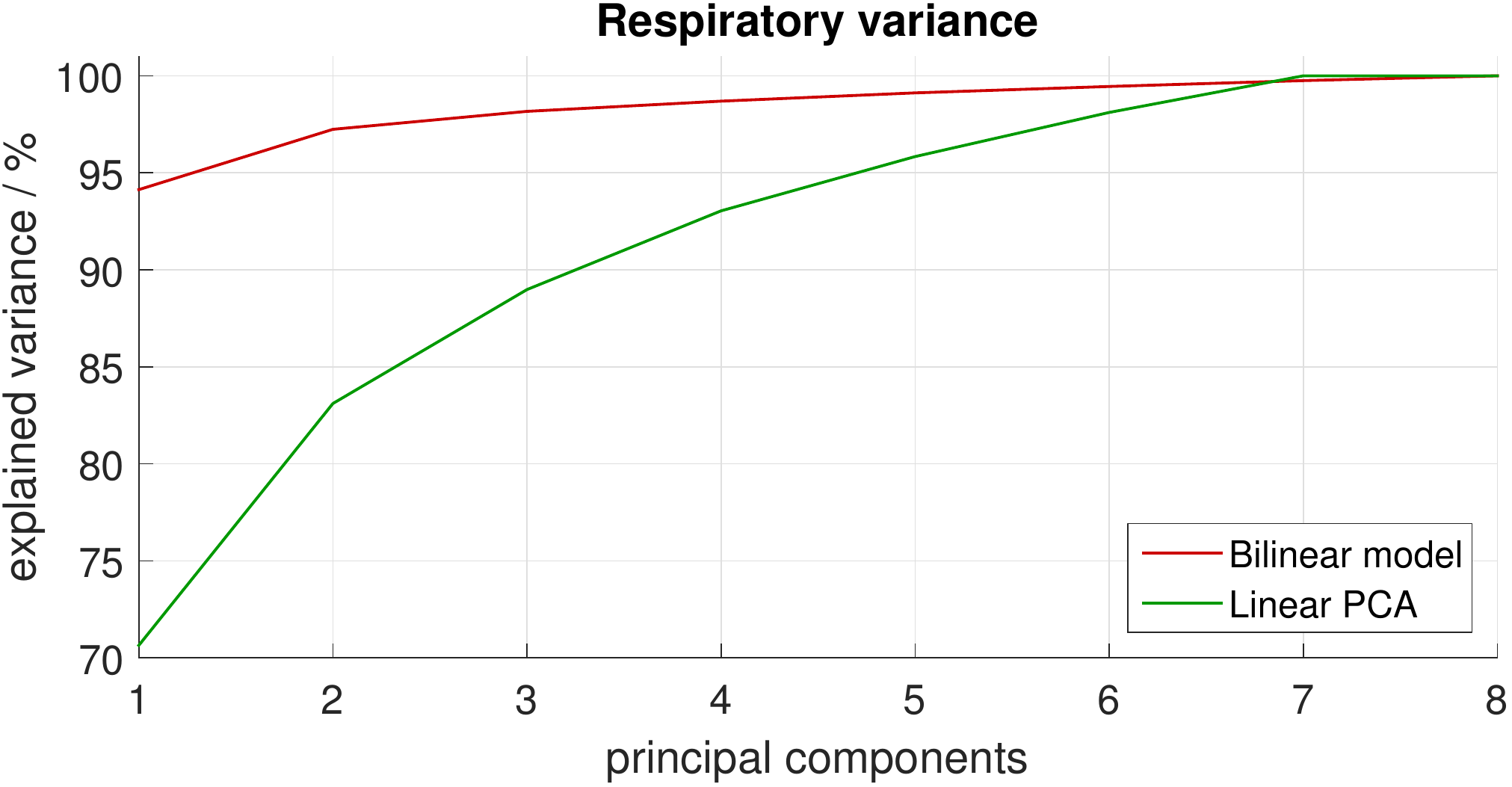}
		}
		
		\centerline{
			\includegraphics[width=0.45\textwidth]{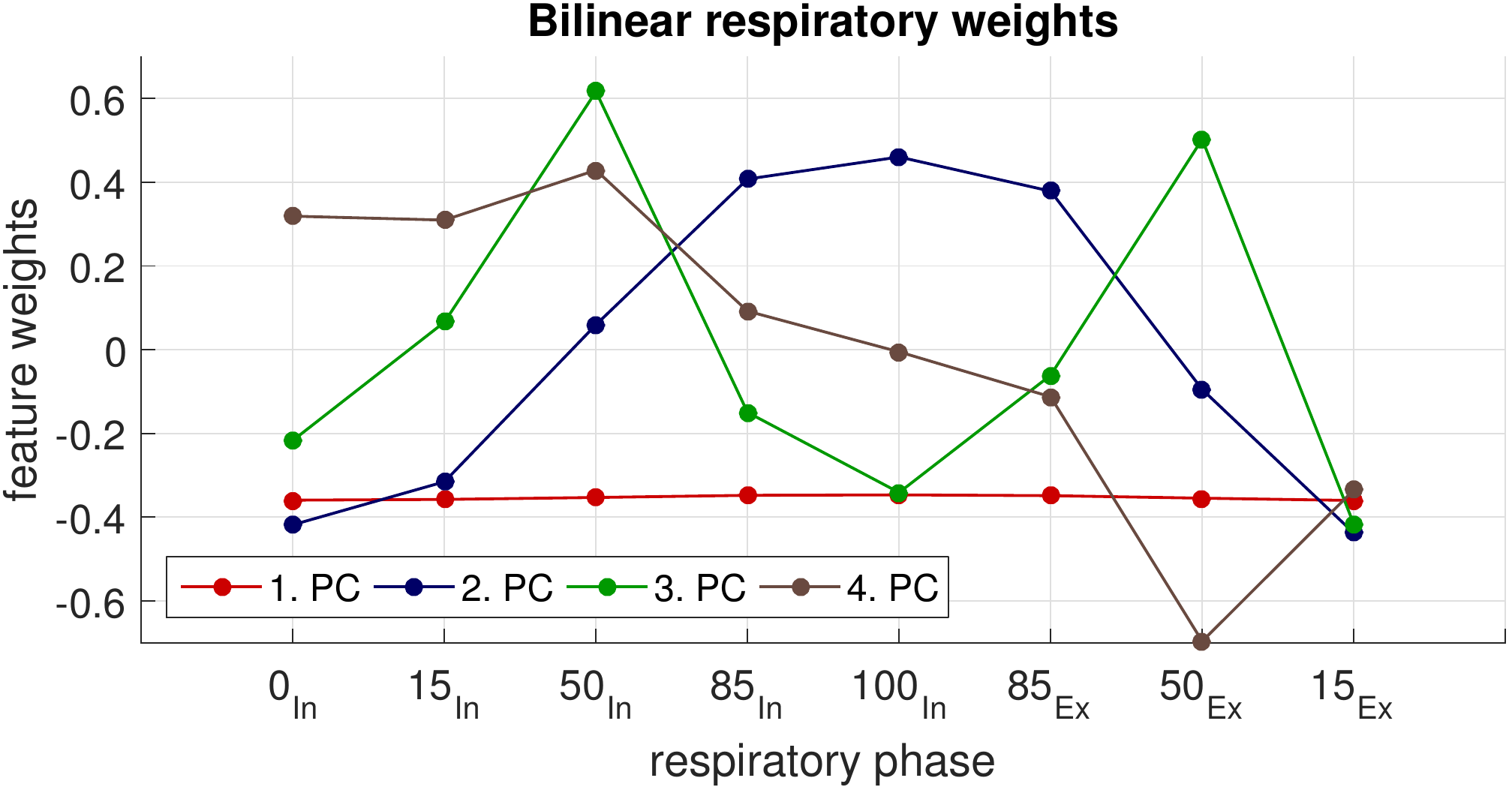}
			\includegraphics[width=0.45\textwidth]{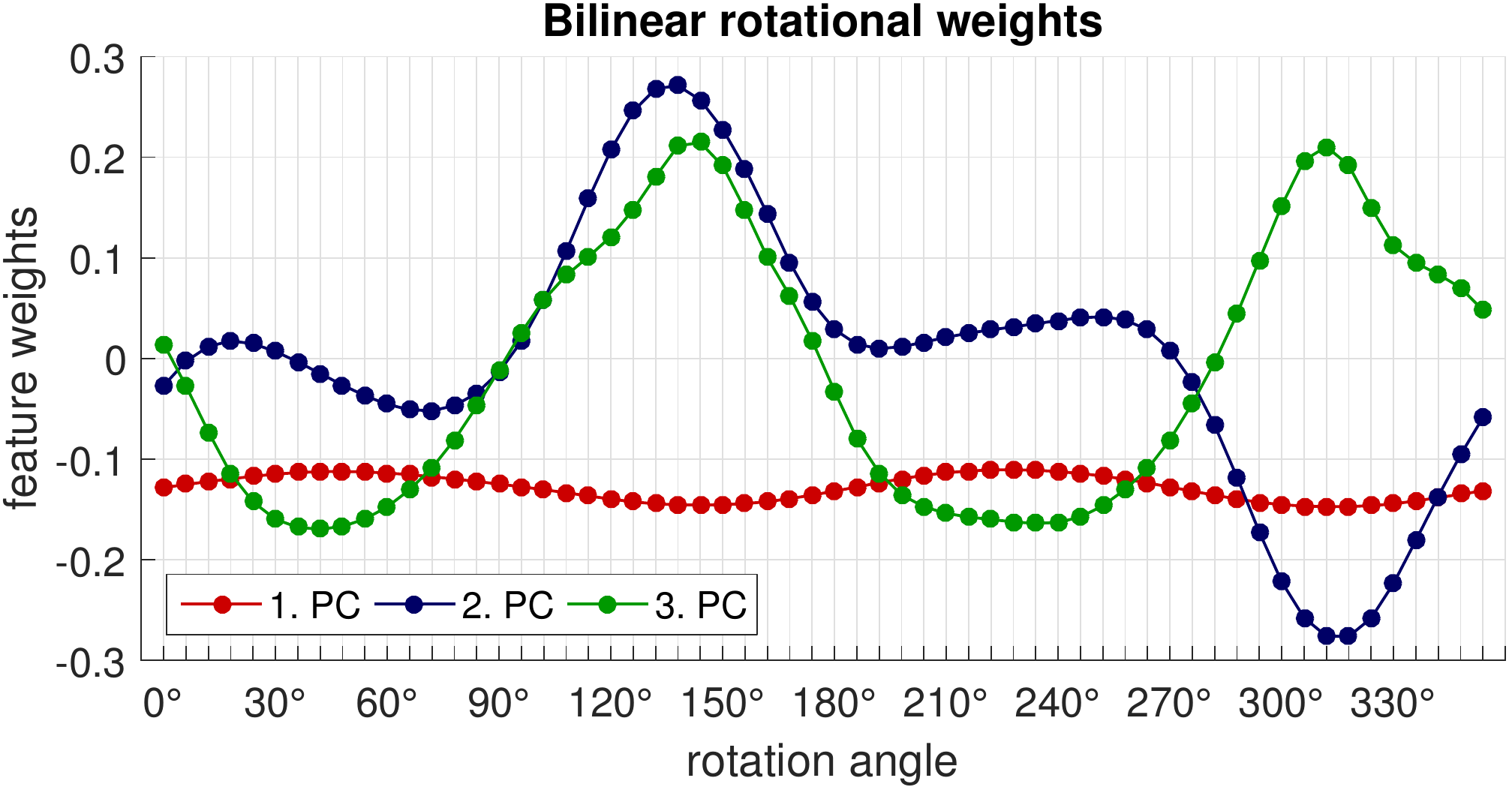}
		}
		\caption{Feature weights for the first few principal components for linear PCA on the 4D CT (t.\ l.), bilinear respiratory (b.\  l.) and rotational weights (b.\ r.) as well as the explained variance caused by respiration in the linear and bilinear case (t.\ r.).}
		
		\label{fig:weightsAndVariance}
	\end{figure}
	\paragraph{Experiment 1.} Fig.\ \ref{fig:weightsAndVariance} shows the weights of the first few principal components for the linear PCA of the volumes as well as each phase and angle in the projection images. Notably, both first bilinear components are near constant. Unlike linear PCA (Eq.~\ref{equ:linear}), data in the bilinear model does not have zero-mean (Eq.~\ref{equ:bilinear}). Consequently, the first component points to the data mean while variation in the respective domain is encoded starting from the second component~\cite{Tenenbaum2000}.
	Appropriately, the $(n+1)$-th bilinear respiratory component corresponds to the $n$-th linear component of the 4D CT indicating that separation of respiratory and angular variance in the projections is in fact achieved.
	\begin{figure}[tb]
		\centering
		\centerline{
			\includegraphics[width=0.45\textwidth]{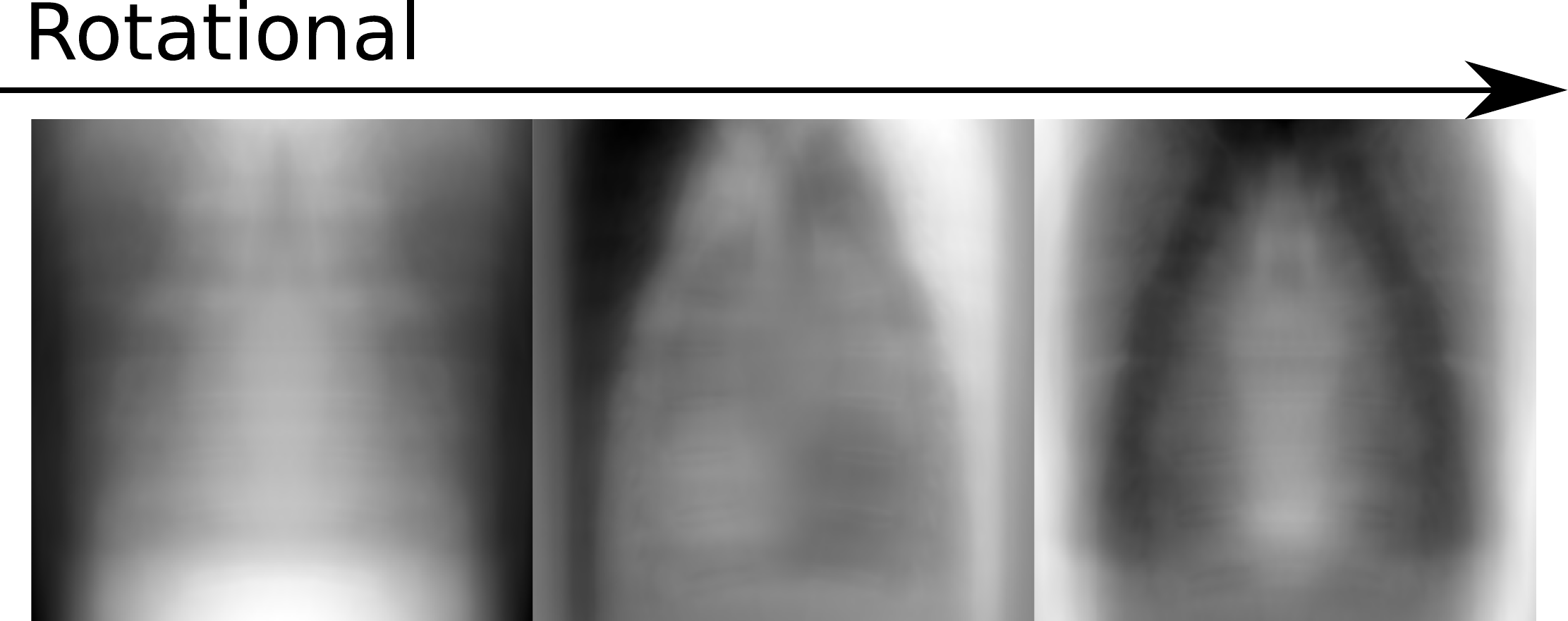}
			$\;\;\;\;\;$
			\includegraphics[width=0.45\textwidth]{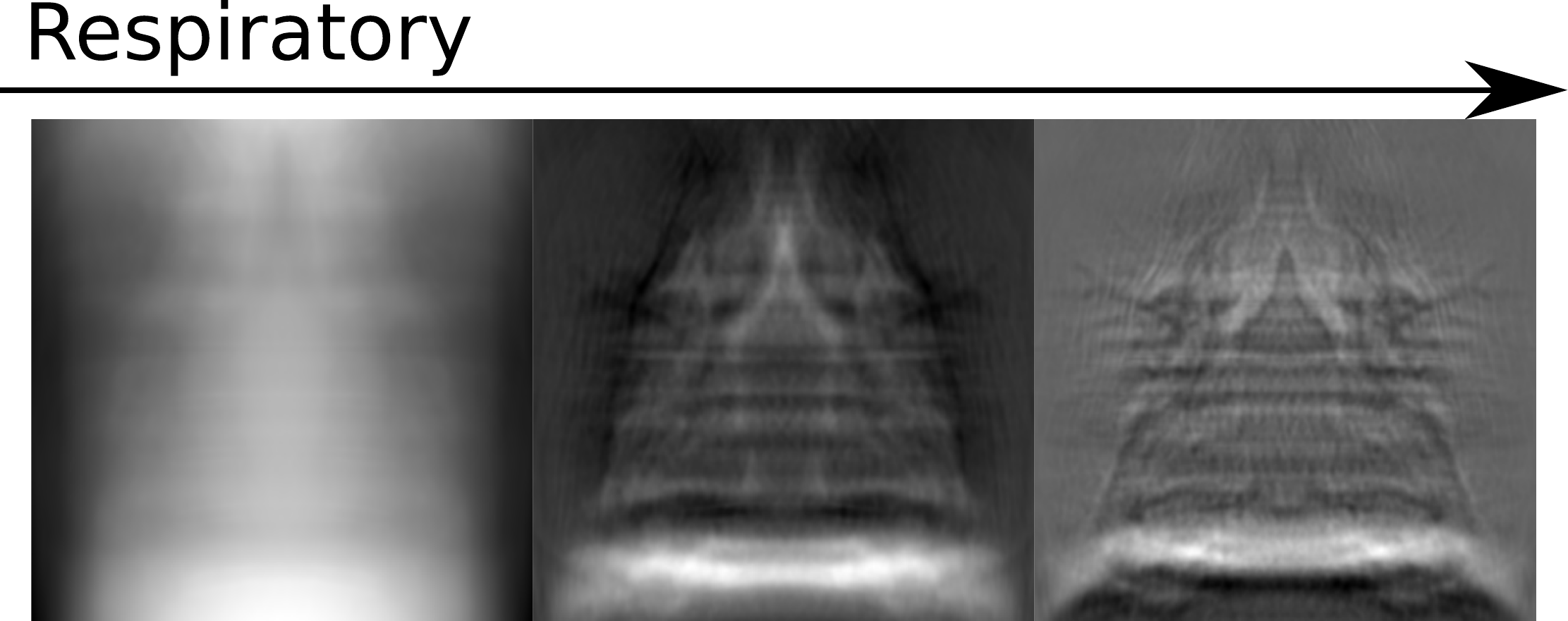}
		}
		\caption{Eigenimages in model tensor $\mathcal{M}$ corresponding to rotational and respiratory features both starting from the same mean.}
		\label{fig:eigenImage}
	\end{figure}
	Additionally, the respiratory variance explained by the principal components is plotted (top right). Most of the variance in the bilinear case is already explained by movement towards the mean. Thus, the linear components better reflect that four components accurately describe over $90\%$ of the data variance.
	As a consequence of the two last mentioned results, one more feature should be extracted than expected by the volumetric 4D SSM, when using the bilinear weights as the driving surrogate. This is also the motivation for the $5 \times 6$ regression matrix chosen in experiment 3.
	Fig.\ \ref{fig:eigenImage} shows the eigenimages corresponding to angular and respiratory variation, respectively. Noticeably, the rotational eigenimages contain mostly low-frequent variation inherent to the moving gantry whereas the respiratory direction encodes comparably high-frequent changes.
	
	% ----------
	% Results 2
	\begin{table}[tb]
		\caption{Mean gray value error and standard deviation for leave-one-out evaluation of representing the projection images in terms of the bilinear model averaged over all phases. Percentage values are given with respect to the mean gray value in the respective reference projection image}
		\label{tab:gray}
		\def\arraystretch{1.2}
		\setlength{\tabcolsep}{0.3em}
		\centering
		\begin{tabular}{| l | r | r | r | r | r | r | r | r | r | r |}
		\hline 
		gray value & \multicolumn{10}{|c|}{trajectory angle} \\ \cline{2-11} 
		error [$\%$] & $18^\circ$&$54^\circ$&$90^\circ$&$126^\circ$&$162^\circ$&$198^\circ$&$234^\circ$&$270^\circ$&$306^\circ$&$342^\circ$\\ \hline
				
		\multirow{2}{*}{Pat 1} &$3.20$&$2.68$&$3.74$&$3.88$&$4.82$&$3.17$&$2.79$&$3.95$&$5.43$&$4.66$\\ 
		&$ \pm 3.41$&$ \pm 2.40$&$ \pm 4.02$&$ \pm 4.00$&$ \pm 4.72$&$ \pm 2.97$&$ \pm 2.53$&$ \pm 3.89$&$ \pm 6.33$&$ \pm 4.78$\\ 
		\hline
		\multirow{2}{*}{Pat 2} &$2.63$&$2.54$&$2.84$&$3.03$&$3.37$&$2.88$&$2.67$&$3.06$&$3.20$&$4.07$\\ 
		&$ \pm 2.41$&$ \pm 2.27$&$ \pm 2.75$&$ \pm 2.80$&$ \pm 3.01$&$ \pm 2.62$&$ \pm 2.38$&$ \pm 2.85$&$ \pm 2.70$&$ \pm 3.54$\\ 
		\hline
		\multirow{2}{*}{Pat 3} &$2.78$&$2.56$&$3.05$&$3.58$&$3.64$&$2.86$&$2.62$&$3.08$&$4.71$&$4.28$\\ 
		&$ \pm 2.64$&$ \pm 2.37$&$ \pm 3.35$&$ \pm 3.87$&$ \pm 3.16$&$ \pm 2.65$&$ \pm 2.39$&$ \pm 2.91$&$ \pm 5.03$&$ \pm 3.83$\\
		\hline
		\multirow{2}{*}{Pat 4} &$1.72$&$1.67$&$1.81$&$2.12$&$1.94$&$1.78$&$1.78$&$1.98$&$2.26$&$2.43$\\ 
		&$ \pm 1.58$&$ \pm 1.51$&$ \pm 1.65$&$ \pm 2.39$&$ \pm 1.72$&$ \pm 1.57$&$ \pm 1.56$&$ \pm 1.77$&$ \pm 2.15$&$ \pm 2.19$\\ 
		\hline
		\multirow{2}{*}{Pat 5} &$2.68$&$2.47$&$2.76$&$2.98$&$2.96$&$2.95$&$2.56$&$2.96$&$3.36$&$3.57$\\ 
		&$ \pm 2.69$&$ \pm 2.35$&$ \pm 2.49$&$ \pm 2.90$&$ \pm 2.61$&$ \pm 3.07$&$ \pm 2.41$&$ \pm 2.69$&$ \pm 3.02$&$ \pm 3.19$\\ 
		\hline
		\end{tabular}
	\end{table}
	\begin{figure}[tb]
		\centering
		\includegraphics[width=0.85\textwidth]{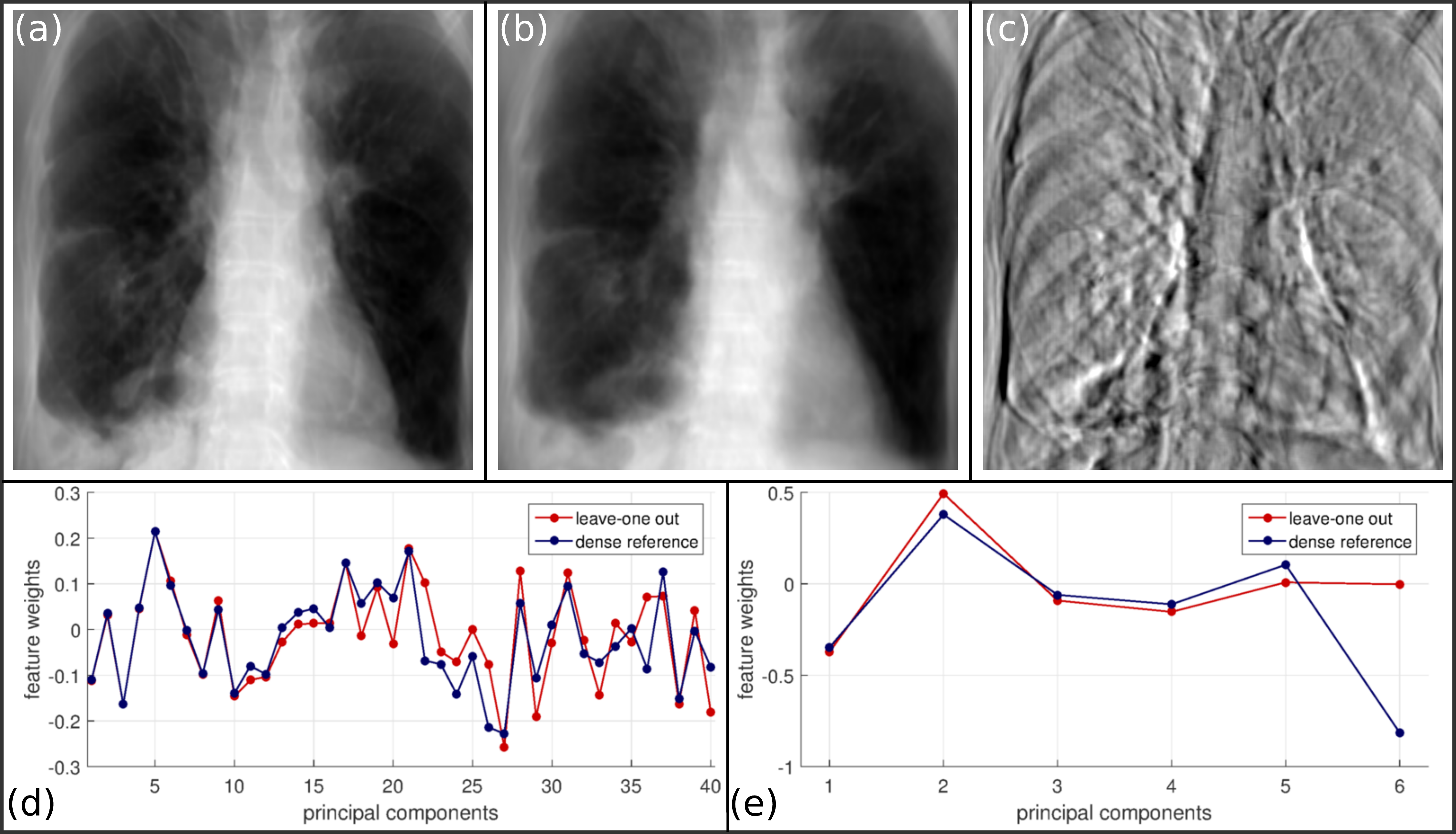}
		\caption{Example reconstruction for $85\%$ exhale phase at $234^\circ$ angle. (a) Original DRR sample. (b) Leave-one-out bilinear reconstruction. (c) Difference image with level/window $-0.15$/$3.75$. (d) Rotational weights from dense bilinear model and interpolated weights for the loo estimation. (e) Respiratory weight estimate. }
		\label{fig:weightEstimation}
	\end{figure}
		
	\paragraph{Experiment 2.} Regarding the bilinear decomposition, Tab.\ \ref{tab:gray} lists the percentage mean gray value error in the reconstructed projection images for each test-angle averaged over all estimated phases. The average error was $1.28 \pm 1.27$ compared to a reference mean gray value of $44.14 \pm 12.39$.
	Exemplarily for one phase-angle combination of patient 1, Fig.\ \ref{fig:weightEstimation} shows the leave-one-out estimation result for $85\%_\text{Ex}$ and $234^\circ$. The proposed B-spline interpolation yields rotational weights close to the dense bilinear model up to the 10th component. Since both sets of weights correspond to slightly different eigenvectors, due to one model being trained on less data, deviation especially in the lower components is to be expected. Still, four to five respiratory weights are estimated accurately which, as shown previously, is sufficient to recover over $90\%$ respiratory variance. As such, we believe they contain much more information than just the respiratory phase.
	
	% ----------
	% Results 3
	\begin{table}[tb]
			\caption{Mean voxel error and standard deviation in HU for each patient and phase averaged over all voxels and angles. The average standard deviation between angles was only  $1.87\,\text{HU}$ within the same patient}
			\label{tab:HU}		
			\def\arraystretch{1.2}
			\setlength{\tabcolsep}{0.3em}
			\centering
			\begin{tabular}{| l | r | r | r | r | r | r | r | r |}
				\hline 
				mean voxel & \multicolumn{8}{|c|}{respiratory phase} \\ \cline{2-9} 
			 	error [HU] & $0_\text{In}$&$15_\text{In}$&$50_\text{In}$&$85_\text{In}$&$100_\text{In}$&$85_\text{Ex}$&$50_\text{Ex}$&$15_\text{Ex}$\\ \hline
			 	
			 	\multirow{2}{*}{Pat 1}  &$21.64$&$17.26$&$22.48$&$15.13$&$16.68$&$16.53$&$23.82$&$21.59$\\ 
										&$ \pm 51.42$&$ \pm 42.38$&$ \pm 51.42$&$ \pm 33.69$&$ \pm 38.87$&$ \pm 38.19$&$ \pm 55.06$&$ \pm 51.87$\\ \hline
				\multirow{2}{*}{Pat 2} &$84.24$&$59.15$&$51.50$&$42.42$&$48.06$&$42.39$&$47.66$&$58.86$\\ 
										&$ \pm 149.73$&$ \pm 94.72$&$ \pm 74.01$&$ \pm 54.16$&$ \pm 65.61$&$ \pm 56.18$&$ \pm 65.68$&$ \pm 86.28$\\ \hline
				\multirow{2}{*}{Pat 3} &$46.47$&$58.57$&$46.56$&$50.69$&$45.38$&$48.80$&$47.60$&$42.25$\\ 
										&$ \pm 64.85$&$ \pm 86.78$&$ \pm 67.06$&$ \pm 74.59$&$ \pm 63.01$&$ \pm 72.00$&$ \pm 68.27$&$ \pm 55.54$\\ \hline
				\multirow{2}{*}{Pat 4} 	&$82.01$&$60.47$&$53.99$&$50.00$&$51.09$&$53.65$&$86.92$&$67.74$\\ 
										&$ \pm 114.07$&$ \pm 69.10$&$ \pm 58.59$&$ \pm 50.49$&$ \pm 52.06$&$ \pm 56.30$&$ \pm 123.08$&$ \pm 77.82$\\ \hline
				\multirow{2}{*}{Pat 5} &$87.08$&$64.21$&$56.32$&$50.17$&$52.17$&$50.91$&$56.74$&$117.47$\\ 
										&$ \pm 133.69$&$ \pm 82.35$&$ \pm 67.89$&$ \pm 57.58$&$ \pm 59.19$&$ \pm 60.12$&$ \pm 68.84$&$ \pm 159.99$\\ \hline
			\end{tabular}
		\end{table}
		
	\begin{figure}[tb]
		\centering
		\includegraphics[width=0.8\textwidth]{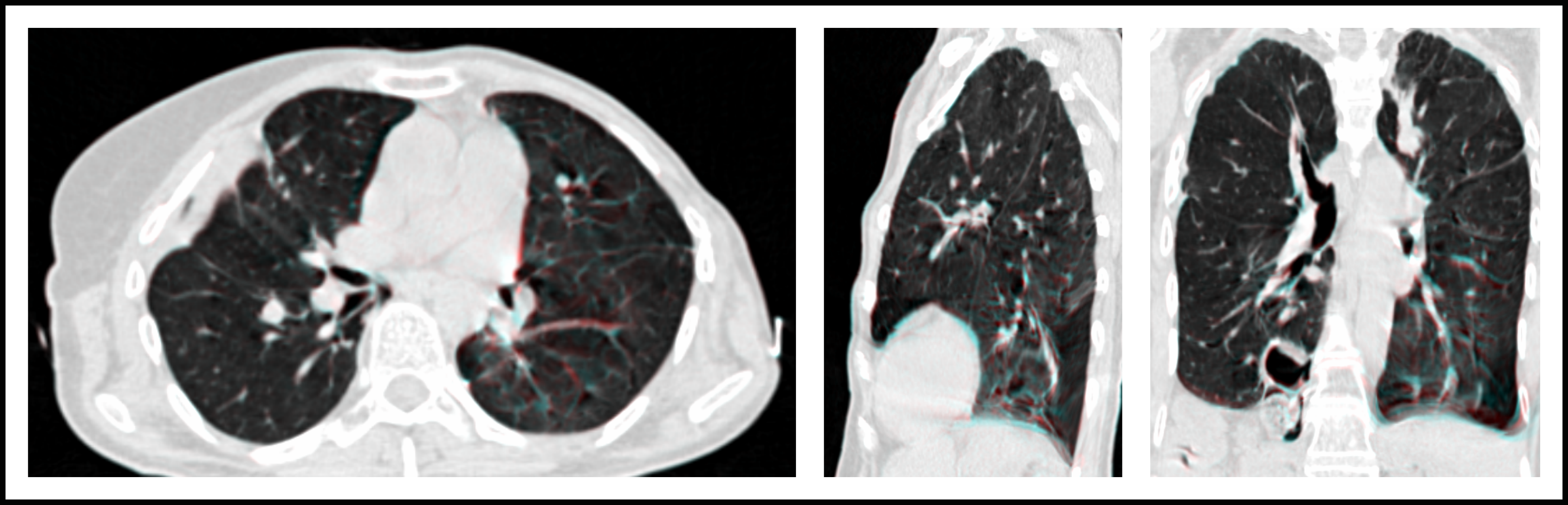}
		\caption{Estimation visualization for patient 1 at $85\%$ exhale phase estimated from projections at $234^\circ$ angle. The overlay displays the original CT in cyan and the estimated volume in red, adding up to gray for equal HU.}
		\label{fig:HU}
	\end{figure}
		
	\paragraph{Experiment 3.} To substantiate this assumption, Tab.\ \ref{tab:HU} provides the mean HU-errors in the estimated CT volumes using projections at the ten test angles for each patient. Errors showed very small deviation w.r.t.\ the acquisition angle, indicating that the respiratory domain is extracted sufficiently regardless of the view. Given an HU range from $-1000$ to $3000$, a mean error of $25$ to $100$ indicates reasonable performance.
	For visualization, Fig.\ \ref{fig:HU} shows the estimated CT for patient 1 at $85\%$ exhale phase, that was estimated using a projection at a trajectory angle of $234^\circ$. Deviation is most prominent at vessel structures within the lung as well as at the left side of the diaphragm.

	In this scenario, the motion representation is covered by a HU-based SSM to generate CT volumes for different respiratory weights. This is of course interchangeable by, for instance, 3D vector fields obtained via deformable image registration. In the context of motion tracking in IGRT, the estimated displacements could then be used to steer the treatment beam according to the tumor motion while at the same time enabling quality assurance in terms of 4D dose verification~\cite{Prasetio2018}.
	However, the main focus of this work was to provide proof-of-concept for the angular-respiratory decoupling process for which a HU-based SSM was sufficient. In future work, we will investigate the potential to predict entire dense deformation fields.

	\paragraph{} Our current leave-one-out evaluation assumed two simplifications, that will pose additional challenges. First, a perfect baseline registration of the training CT to the projection images may not be the case in every scenario. However, for the case of radiation therapy accurate alignment of patient and system is a prerequisite for optimal treatment.
	Second, no anatomical changes between the 4D CT and the rotational scan are taken into account. Further investigation on how these effects interfere with the decomposition are subject to future work.
	
	\section{Conclusion}
	\label{sec:conclusion}
	In this paper, we demonstrate that the X-ray transform under respiratory and angular variation can be expressed in terms of a bilinear model given a continuous trajectory and that motion is observable in every projection. Using a prior 4D CT, we show that projection images on the trajectory can be bilinearly decomposed into rotational and respiratory components. Prior knowledge about the gantry angle is used to solve this ill-posed out-of-sample problem. Results for both 2D DRRs and estimated 3D volumes demonstrate that up to five components of the respiratory variance are recovered independent of the view-angle. These explain more than $90\,\%$ of the volumetric variation. As such, recovery of 3D motion seems possible. Currently our study is limited by two simplifications, namely perfect alignment and no inter-acquisition changes. Their investigation is subject to future work.

	\section*{Acknowledgement}
	\label{sec:acknowledgement}
	This work was partially conducted at the ACRF Image X Institute as part of a visiting research scholar program. The authors gratefully acknowledge funding of this research stay by the Erlangen Graduate School in Advanced Optical Technologies (SAOT).

	\bibliographystyle{splncs04}
	\bibliography{0133}
\end{document}